\documentclass[10pt,twocolumn,letterpaper]{article}

\usepackage[pagenumbers]{cvpr} 

\usepackage{graphicx}
\usepackage{amsmath}
\usepackage{amssymb}
\usepackage{booktabs}
\usepackage{multirow}
\usepackage[para]{footmisc}

\usepackage[pagebackref,breaklinks,colorlinks]{hyperref}

\usepackage[capitalize]{cleveref}
\crefname{section}{Sec.}{Secs.}
\Crefname{section}{Section}{Sections}
\Crefname{table}{Table}{Tables}
\crefname{table}{Tab.}{Tabs.}

\def\cvprPaperID{2958} 
\def\confName{CVPR}
\def\confYear{2023}

\begin{document}



\title{BUOL: A Bottom-Up Framework with Occupancy-aware Lifting for \\ Panoptic 3D Scene Reconstruction From A Single Image}

\author{Tao Chu$^{1, 2}$\footnotemark \hspace{0.1em}
	, Pan Zhang$^2$, Qiong Liu$^1$\footnotemark \hspace{0.3em}, 
    Jiaqi Wang$^2$
	\\ 
	$^1$ South China University of Technology \quad
	$^2$ Shanghai AI Laboratory
    \\
    {\tt\small \{chutao, zhangpan, wangjiaqi\}@pjlab.org.cn \quad
    liuqiong@scut.edu.cn}
}

\maketitle
\renewcommand{\thefootnote}{\fnsymbol{footnote}}
\footnotetext[1]{\hspace{-0.5em}Intern at Shanghai AI Laboratory.}
\footnotetext[2]{\hspace{-0.5em}Corresponding author.}

\begin{abstract}

Understanding and modeling the 3D scene from a single image is a practical problem. A recent advance proposes a panoptic 3D scene reconstruction task that performs both 3D reconstruction and 3D panoptic segmentation from a single image. Although having made substantial progress, recent works only focus on top-down approaches that fill 2D instances into 3D voxels according to estimated depth, which hinders their performance by two ambiguities. (1) \textbf{instance-channel ambiguity}: The variable ids of instances in each scene lead to ambiguity during filling voxel channels with 2D information, confusing the following 3D refinement. (2) \textbf{voxel-reconstruction ambiguity}: 2D-to-3D lifting with estimated single view depth only propagates 2D information onto the surface of 3D regions, leading to ambiguity during the reconstruction of regions behind the frontal view surface. In this paper, we propose \textbf{BUOL}, a \textbf{B}ottom-\textbf{U}p framework with \textbf{O}ccupancy-aware \textbf{L}ifting to address the two issues for panoptic 3D scene reconstruction from a single image. For \textbf{instance-channel ambiguity}, a bottom-up framework lifts 2D information to 3D voxels based on deterministic semantic assignments rather than arbitrary instance id assignments. The 3D voxels are then refined and grouped into 3D instances according to the predicted 2D instance centers. For \textbf{voxel-reconstruction ambiguity}, the estimated multi-plane occupancy is leveraged together with depth to fill the whole regions of things and stuff. Our method shows a tremendous performance advantage over state-of-the-art methods
on synthetic dataset 3D-Front and real-world dataset Matterport3D. Code and models are available in \href{https://github.com/chtsy/buol}{https://github.com/chtsy/buol}.

\begin{figure}[t!]
\centering
\small
\setlength\tabcolsep{0pt}
\includegraphics[width=0.9\columnwidth]{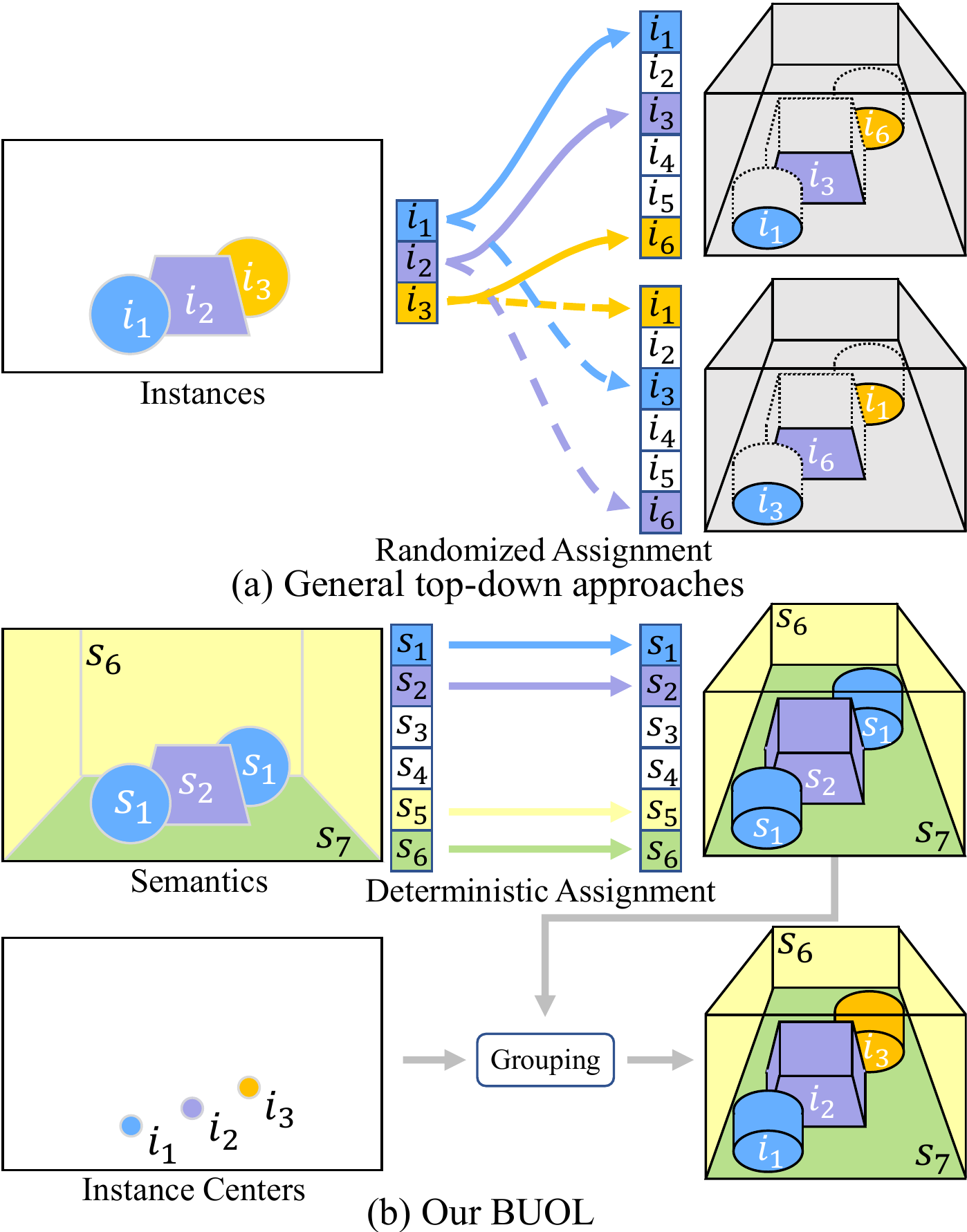}
\caption{\textbf{Comparison of the feature lifting from 2D to 3D.} \textbf{(a) General Top-down approaches:} Feature lifting by depth with the two randomized instance assignments in the top-down framework.
The predicted 2D instance masks $\{i_1, i_2,  i_3\}$ are lifted to only the surface of 3D instances at variable channels, such as $\{i_1, i_3, i_6\}$ or $\{i_3, i_6, i_1\}$, which results in instance-channel ambiguity and voxel-reconstruction ambiguity. \textbf{(b) Our BUOL:} Occupancy-aware lifting with the deterministic semantic assignment in the bottom-up framework. The predicted 2D semantic category maps $\{s_1, s_2, s_6, s_7\}$ are lifted to the whole regions of things ($s_1, s_2$) and stuff ($s_6, s_7$), and the voxels are finally grouped into 3D instances $\{i_1, i_2, i_3\}$ by corresponding 2D instance centers.}
\vspace{-6pt}
\label{fig:teaser}
\end{figure}

\end{abstract}

\section{Introduction}


Joint learning of 3D reconstruction and perception is a challenging and practical problem for various applications. Existing works focus on combining 3D reconstruction with semantic segmentation~\cite{murez2020atlas,li2020anisotropic} or instance segmentation~\cite{gkioxari2019mesh,kuo2020mask2cad,nie2020total3dunderstanding}. Recently, a pioneer work~\cite{dahnert2021panoptic} unifies the tasks of 3D reconstruction, 3D semantic segmentation, and 3D instance segmentation into panoptic 3D scene reconstruction from a single RGB image, which assigns a category label (i.e. a thing category with easily distinguishable edges, such as tables, or a stuff category with indistinguishable edges, such as wall)~\cite{kirillov2019panoptic} and an instance id (if the voxel belongs to a thing category) to each voxel in the 3D volume of the camera frustum. 

Dahnert et al. \cite{dahnert2021panoptic} achieve this goal in a top-down pipeline that lifts 2D instance masks to channels of 3D voxels and predicts the panoptic 3D scene reconstruction in the following 3D refinement stage. Their method first estimates 2D instance masks and the depth map. The 2D instance masks are then lifted to fill voxel channels on the front-view surface of 3D objects using the depth map. Finally, a 3D model is adopted to refine the lifted 3D surface masks and attain panoptic 3D scene reconstruction results of all voxels.

After revisiting the top-down panoptic 3D scene reconstruction framework, we find two crucial limitations which hinder its performance, as shown in Figure~\ref{fig:teaser}(a). First, \textbf{instance-channel ambiguity}: the number of instances varies in different scenes. Thus lifting 2D instance masks to fill voxel channels can not be achieved by a deterministic instance-channel mapping function. Dahnert et al. \cite{dahnert2021panoptic} propose to utilize a randomized assignment that randomly assigns instance ids to the different channels of voxel features. For example, two possible random assignments are shown in Figure.~\ref{fig:teaser}(a), where solid and dashed arrow lines with the same color indicate a 2D mask is assigned to different voxel feature channels. This operator leads to instance-channel ambiguity, where an instance id may be assigned to an arbitrary channel, confusing the 3D refinement model. In addition, we experimentally discuss the impact of different instance assignments (\eg, random or sorted by category) on performance in Section \ref{sec:experiments}.
Second, \textbf{voxel reconstruction ambiguity}: 2D-to-3D lifting with depth from a single view can only propagate 2D information onto the frontal surface in the camera frustum, causing ambiguity during the reconstruction of regions behind the frontal surface. As shown by dashed black lines in the right of Figure~\ref{fig:teaser}(a), the 2D information is only propagated to the frontal surface of initialized 3D instance masks, which is challenging for 3D refinement model to reconstruct the object regions behind the frontal surface accurately.

In this paper, we propose \textbf{BUOL}, a \textbf{B}ottom-\textbf{U}p framework with \textbf{O}ccupancy-aware \textbf{L}ifting to address the above two ambiguities for panoptic 3D scene reconstruction from a single image. For instance-channel ambiguity, our bottom-up framework lifts 2D semantics to 3D semantic voxels, as shown in Figure.~\ref{fig:teaser}(b). Compared to the top-down methods shown in Figure.~\ref{fig:teaser}(a), instance-channel ambiguity is tackled by a simple deterministic assignment mapping from semantic category ids to voxel channels. The voxels are then grouped into 3D instances according to the predicted 2D instance centers. For voxel-reconstruction ambiguity, as shown in Figure.~\ref{fig:teaser}(b), the estimated multi-plane occupancy is leveraged together with depth by our occupancy-aware lifting mechanism to fill regions inside the things and stuff besides front-view surfaces for accurate 3D refinement.

Specifically, our framework comprises a 2D priors stage, a 2D-to-3D lifting stage, and a 3D refinement stage.
In the 2D priors stage, the 2D model predicts 2D semantic map, 2D instance centers, depth map, and multi-plane occupancy. The multi-plane occupancy presents whether the plane at different depths is occupied by 3D things or stuff. In the 2D-to-3D lifting stage, leveraging estimated multi-plane occupancy and depth map, we lift 2D semantics into deterministic channels of 3D voxel features inside the things and stuff besides the front-view surfaces. 
In the 3D refinement stage, we predict dense 3D occupancy in each voxel for reconstruction. Meanwhile, the 3D semantic segmentation is predicted for both the thing and stuff categories. The 3D offsets towards the 2D instance centers are also estimated to identify voxels belonging to 3D objects.
The ground truth annotations of 3D panoptic reconstruction, \ie, 3D instance/semantic segmentation masks and dense 3D occupancy, 
can be readily converted to 2D instance center, 2D semantic segmentation, depth map, multi-plane occupancy, and 3D offsets for our 2D and 3D supervised learning.
During inference, we assign instance ids to 3D voxels occupied by thing objects based on 2D instance centers and 3D offsets, attaining final panoptic 3D scene reconstruction results.

Extensive experiments show that the proposed bottom-up framework with occupancy-aware lifting outperforms prior competitive approaches.
On the pre-processed 3D-Front \cite{fu20213d} and Matterport3D \cite{chang2017matterport3d}, our method achieves +11.81\% and +7.46\% PRQ (panoptic reconstruction quality) over the state-of-the-art method~\cite{dahnert2021panoptic}, respectively.


\section{Related Work}


\noindent\textbf{3D reconstruction.}
Single-view 3D reconstruction learns 3D geometry from a single-view image.
Pixel2Mesh attempts to progressively deform an initialized ellipsoid mesh for a single object, while DISN predicts the underlying signed distance fields to generate the single 3D mesh.
UCLID-Net~\cite{guillard2020uclid} back-projects 2D features by the regressed depth map to object-aligned 3D feature grids, and
CoReNet~\cite{popov2020corenet} is proposed to lift 2D features to 3D volume by ray-traced skip connections.
To reconstruct the object or scene in more detail, some works adopt multi-view images as input.
Pix2Vox~\cite{xie2019pix2vox} is proposed to select high-quality reconstructions for each part in 3D volumes generated by different view images.
TransformerFusion~\cite{bozic2021transformerfusion} also selectively stores features extracted from multi-view images.


\noindent\textbf{3D segmentation.}
Some 3D segmentation methods directly use a basic geometry as input. For 3D semantic segmentation,
3DMV~\cite{dai20183dmv} combines the features extracted from 3D geometry with lifted multi-view image features to predict per-voxel semantics.
ScanComplete~\cite{dai2018scancomplete} is proposed to predict complete 3D geometry with per-voxel semantics by devising 3D geometry with filter kernels invariant to the overall scene size.

For 3D instance segmentation, there exist some top-down and bottom-up methods as follows.
Some methods~\cite{hou20193d, hou2020revealnet} based on box proposals predicted by 3D-RPN pay more attention to the fusion of 3D geometry features and lifted image features.
SGPN~\cite{wang2018sgpn} predicts point grouping proposals for point cloud instance segmentation. RfD-Net~\cite{nie2021rfd} focuses on predicting instance mesh of the high objectness proposal predicted by point cloud proposal network. Instead of directly regressing bounding box, GSPN~\cite{yi2019gspn} generates proposals by reconstructing shapes from noisy observations to provide location of instances.

Most bottom-up methods adopt center as the goal of instance grouping.
PointGroup~\cite{jiang2020pointgroup} and TGNN~\cite{huang2021text} learn to extract per-point features and predict offsets to shift each point toward its object center.
Lahoud et al. \cite{lahoud20193d} propose to generate instance labels by learning a metric that groups parts of the same object instance and estimates the direction toward the instance’s center of mass.
There also exist other bottom-up methods.
OccuSeg~\cite{han2020occuseg} predicts the number of occupied voxels for each instance to guide the clustering stage of 3D instance segmentation.
HAIS~\cite{chen2021hierarchical} introduces point aggregation for preliminarily clustering points to sets and set aggregation for generating complete instances.


\noindent\textbf{3D segmentation with reconstruction.}
For 3D semantic segmentation with reconstruction,
Atlas~\cite{murez2020atlas} is proposed to directly regress a truncated signed distance function (TSDF) from a set of posed RGB images for jointly predicting the 3D semantic segmentation of the scene.
AIC-Net~\cite{li2020anisotropic} is proposed to apply anisotropic convolution to the 3D features lifted from 2D features by the corresponding depth to adapt to the dimensional anisotropy property voxel-wisely.


As far as we know, the instance segmentation with reconstruction works follow the top-down pipeline.
Mesh R-CNN~\cite{gkioxari2019mesh} augments Mask R-CNN~\cite{he2017mask} with a mesh prediction branch to refine the meshes converted by predicted coarse voxel representations.
Mask2CAD~\cite{kuo2020mask2cad} and Patch2CAD~\cite{kuo2021patch2cad} leverage the CAD model to match each detected object and its patches, respectively.
Total3DUnderstanding~\cite{nie2020total3dunderstanding} is proposed to prune mesh edges with a density-aware topology modifier to approximate the target shape.
Panoptic 3D Scene Reconstruction from a single image is first proposed by Dahnert et al. \cite{dahnert2021panoptic}, and they deliver a state-of-the-art top-down strategy with Mask R-CNN~\cite{he2017mask} as 2D instance segmentation and random assignment for instance lifting. Our BUOL is the first bottom-up method for panoptic/instance segmentation with reconstruction from a single image.

\section{Methodology}

\begin{figure*}[t!]
\centering
\small
\setlength\tabcolsep{0pt}
\includegraphics[width=1.0\linewidth]{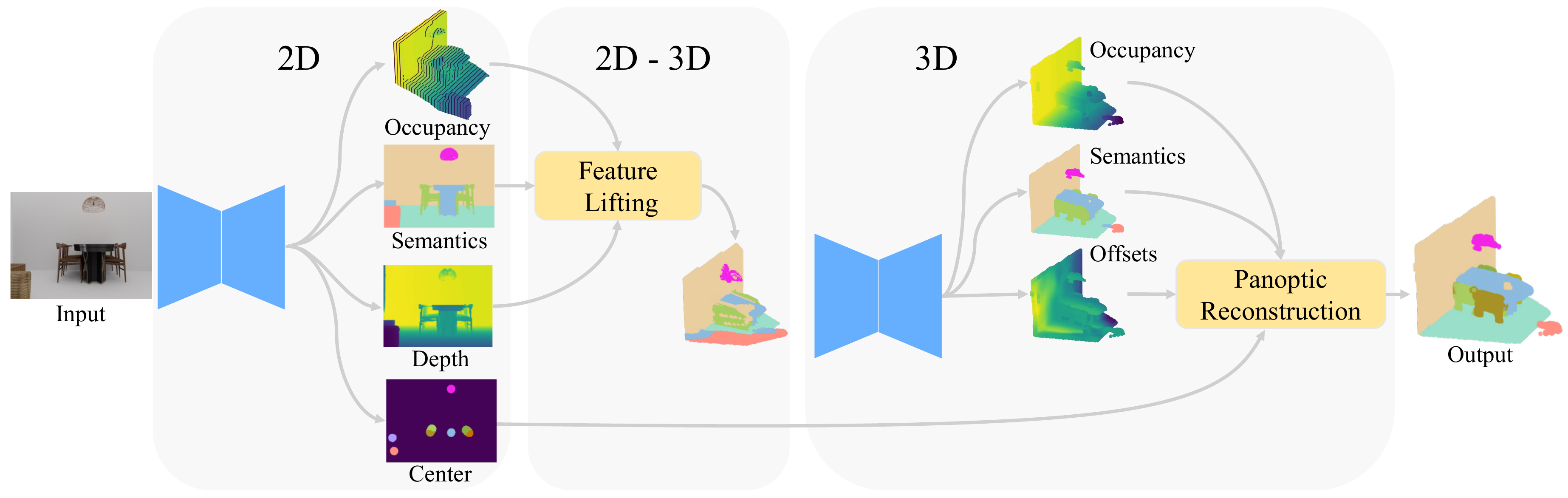}
\vspace{-10pt}
\caption{\textbf{The illustration of our framework.} Given a single image, we first predict 2D priors by 2D model, then lift 2D priors to 3D voxels by our occupancy-aware lifting, and finally predict 3D results using the 3D model and obtain panoptic 3D scene reconstruction results in a bottom-up manner.}
\vspace{-5pt}
\label{fig:framework}
\end{figure*}

In this section, we propose a bottom-up panoptic 3D scene reconstruction method with occupancy-aware lifting. Given a single 2D image, we aim to learn corresponding 3D occupancy and 3D panoptic segmentation. To achieve this goal, as shown in Figure.~\ref{fig:framework}, we first extract the 2D priors, which includes 2D semantics, 2D instance centers,
scene depth, and multi-plane occupancy. Then, an efficient occupancy-aware feature lifting block is designed to lift the 2D priors to 3D features, thus giving a good initialization for the following learning. Finally, a bottom-up panoptic 3D scene reconstruction model is utilized to learn the 3D occupancy and 3D panoptic segmentation, where a 3D refinement model maps the lifted 3D features to 3D  occupancy, 3D semantics, and 3D offsets, and an instance grouping block is designed for 3D panoptic segmentation. In addition, the ground truth of 2D priors and 3D offsets adopted by our method can be easily obtained by ground truth annotations of 3D panoptic reconstruction (\ie 3D semantic map, instance masks, and occupancy).


\subsection{2D Priors Learning}
Given a 2D image $x \in \mathbb{R}^{H\times W \times 3} $, 
where $H$ and $W$ is image height and width, panoptic 3D scene reconstruction aims to map it to semantic labels $\hat{s}^{3d}$ and instance ids $\hat{i}^{3d}$. It's hard to directly learn 3D knowledge from a single 2D image, so we apply a 2D model $F_{\theta}$ to learn rich 2D priors:
\vspace{-3pt}
\begin{equation}
\vspace{-3pt}
   s^{2d}, d, c^{2d},
   o^{mp} = F_{\theta}(x),
\end{equation}
where $s^{2d} \in [0,1]^{H\times W \times C} $ is 2D semantics with $C$ categories.
$d \in \mathbb{R}^{H\times W}$ is the depth map.
$c^{2d} \in \mathbb{R}^{N\times 3}$ is predicted locations of $N$ instance centers ($\mathbb{R}^{N\times 2}$) with corresponding category labels ($\{0,1,...C-1\}^{N\times 1}$).
$o^{mp} \in [0,1]^{H\times W \times M}$ is the estimated multi-plane occupancy which presents whether the $M$ planes at different depths are occupied by 3D things or stuff, and the default $M$ is set as 128.

\subsection{Occupancy-aware Feature Lifting}
\begin{figure}[t!]
\centering
\small
\setlength\tabcolsep{0pt}
\includegraphics[width=0.9\columnwidth]{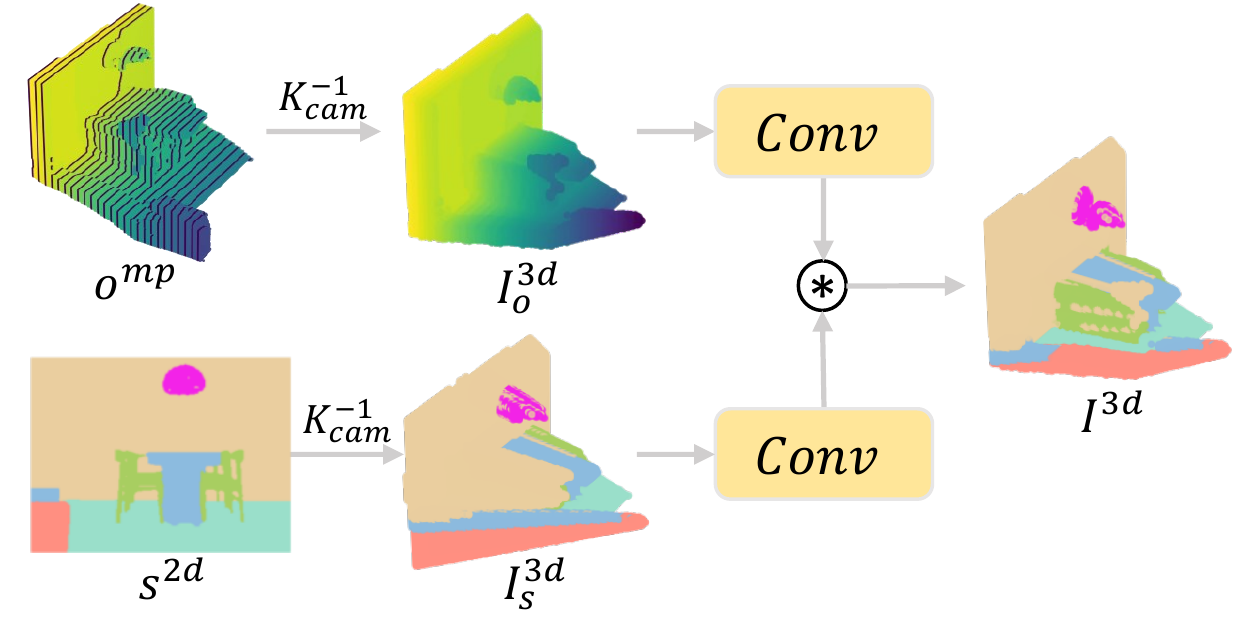}
\vspace{-5pt}
\caption{\textbf{Occupancy-aware Lifting.} We lift multi-plane occupancy and 2D semantics predicted by the 2D model to 3D features. $\ast$ is Hadamard product.}
\vspace{-5pt}
\label{fig:feature_lifting}
\end{figure}

After obtaining the learned 2D priors, we need to lift them to 3D features for the following training. Here, an occupancy-aware feature lifting block is designed for this goal, as shown in Figure.~\ref{fig:feature_lifting}. First, we lift the 2D semantics $s^{2d}$ to coarse 3D semantics $I^{3d}_s$ in the whole region of things and stuff rather than only on the front-view surface adopted by previous work~\cite{dahnert2021panoptic},
\vspace{-3pt}
\begin{equation}
\vspace{-3pt}
    I^{3d}_s(u,v,z) =
    \begin{cases}
    s^{2d}(K_{cam}^{-1}[u,v,1]), & \text{if }\ z \geq d(u,v)
    \\
    0, & \text{otherwise}
    \end{cases}
\end{equation}
where $K_{cam}$ is the camera intrinsic matrix, $d(u,v)$ is depth at location $(u,v)$. The region $z < d(u,v)$ is free space, where is set 0 to ignore.


Then, we resort to multi-plane occupancy $o^{mp}$ learned in the 2D stage to remove the meaningless region of the coarse 3D semantics $I^{3d}_s$ and obtain the lifted 3D features.
Formally, the lifted 3D features are calculated as the product of $I^{3d}_s$ and coarse 3D occupancy $I^{3d}_o$,
\vspace{-3pt}
\begin{equation}
\vspace{-1pt}
\begin{aligned}
    I^{3d} = & Conv(I^{3d}_s) \ast Conv(I^{3d}_o), \text{where} \\ 
    I^{3d}_o(u,v,z) = &
    \begin{cases}
    o^{mp}(K_{cam}^{-1}[u,v,z]), & \text{if }\ z \geq d(u,v)\\
    0, & \text{otherwise}
    \end{cases}
\end{aligned}
\end{equation}
where $Conv$ is a Conv-BN-ReLU block. $\ast$ is Hadamard product.

As shown in Figure.~\ref{fig:feature_lifting}, our multi-plane occupancy can give supplementary shape cues for the occluded region, thus the lifted features are capable to serve as a good 3D initialization for the following 3D refinement, greatly reducing the pressure of the 3D refinement model.

\subsection{Bottom-up Panoptic Reconstruction}
Usually, the lifted 3D features are coarse and cannot be used for panoptic reconstruction directly. To refine the coarse features, a powerful 3D encoder-decoder model $G_{\phi}$ is used to predict 3D occupancy, 3D semantic map, and 3D offsets:
\vspace{-3pt}
\begin{equation}
\vspace{-3pt}
   s^{3d'}, \triangle c^{3d'}, o^{3d} = G_{\phi}(I^3),
\end{equation}
where $s^{3d'}, \triangle c^{3d'}, o^{3d}$ is refined 3D semantic map, 3D offsets and 3D occupancy, respectively.

\begin{figure}[t!]
\centering
\small
\setlength\tabcolsep{0pt}
\includegraphics[width=1.\columnwidth]{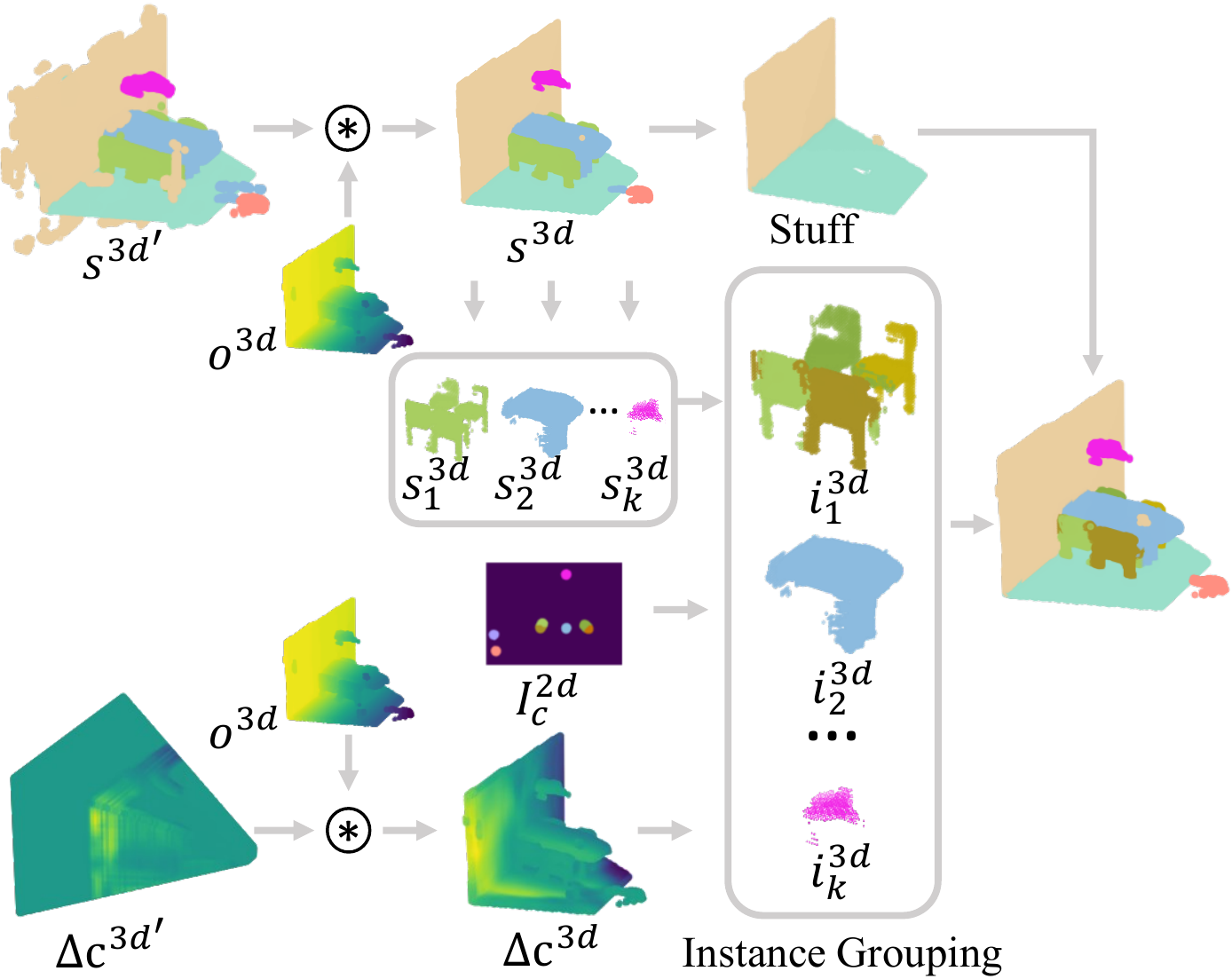}
\vspace{-5pt}
\caption{\textbf{Panoptic Reconstruction.} The predicted 3D semantics and 3D offsets are first refined by 3D occupancy, and then the reconstructed 3D results are combined with 2D instance centers for 3D instance grouping, and finally, 3D instances and stuff are combined to obtain panoptic 3D scene reconstruction. $\ast$ is Hadamard product.}
\vspace{-5pt}
\label{fig:panoptic_reconstrucion}
\end{figure}

The panoptic reconstruction utilizes the refined 3D results for 3D reconstruction and 3D panoptic segmentation, as shown in Figure.~\ref{fig:panoptic_reconstrucion}. For 3D reconstruction, guided by the 3D occupancy $o^{3d}$, we can obtain reconstructed semantics by $s^{3d} = s^{3d'} \ast o^{3d}$ and reconstructed offsets by $\triangle c^{3d} = \triangle c^{3d'} \ast o^{3d}$, where $\ast$ is Hadamard product.
For 3D panoptic segmentation, we need to assign the instance ids to the voxels of the \textit{things}. To achieve this, we propose grouping instances with the estimated 2D instance centers, 3D offsets, and 3D semantics.

\begin{figure}[t!]
\centering
\small
\setlength\tabcolsep{0pt}
\includegraphics[width=1.0\columnwidth]{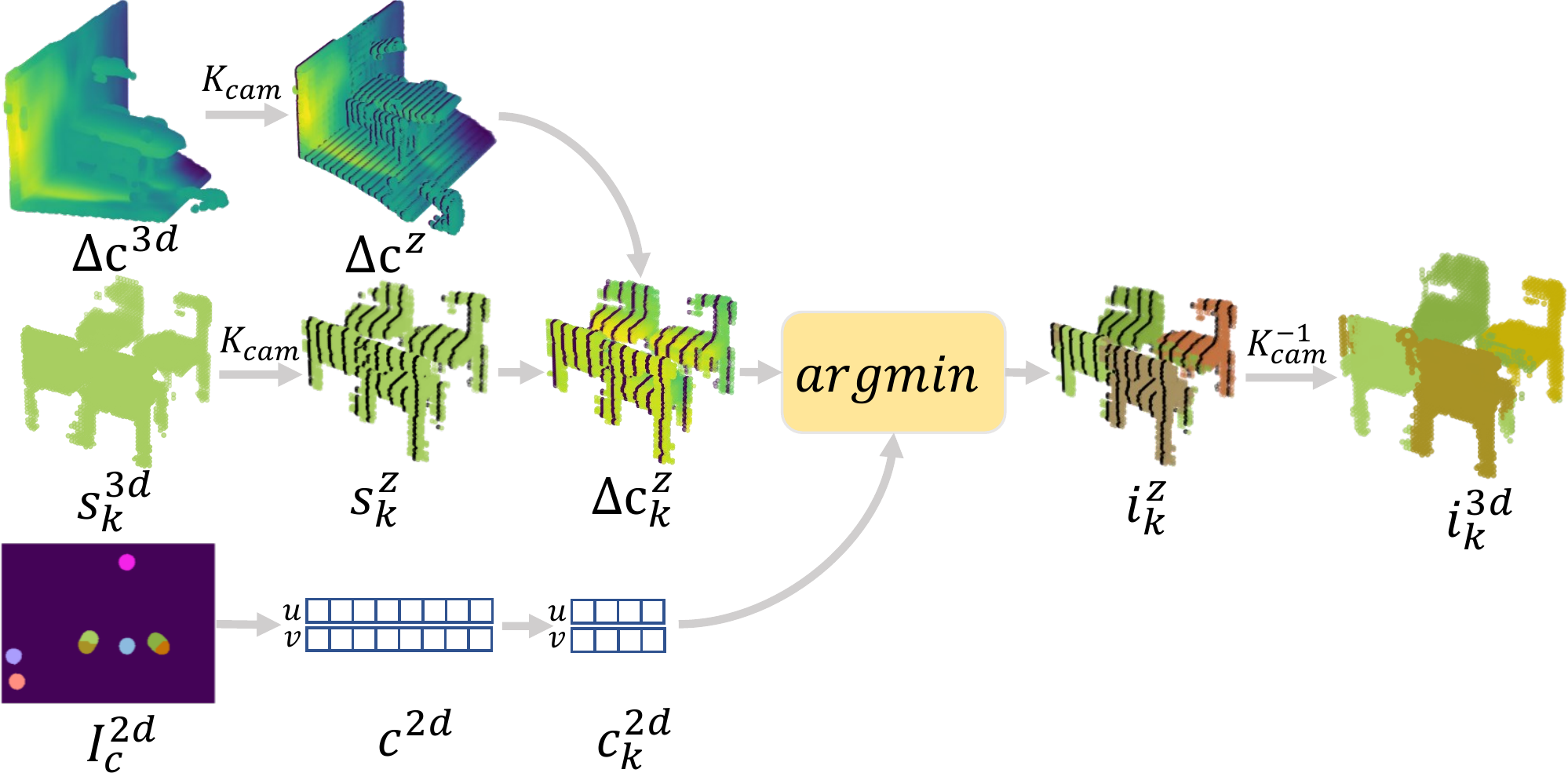}
\vspace{-10pt}
\caption{\textbf{Instance Grouping.} We convert both 2D instance centers and 3D offsets of each category at multi-plane to group 3D instances.}
\vspace{-5pt}
\label{fig:instance_grouping}
\end{figure}

The proposed instance grouping block is shown in Figure.~\ref{fig:instance_grouping}. We first convert 3D offsets $\triangle c^{3d}$ to multi-plane by $ \triangle c^{z} = \triangle c^{3d}(K_{cam}[u,v,z]) $, where $z \in \{0,1,...,M-1\}$ corresponds to different depths. Then multi-plane semantics of category $k$ can also be calculated by $s_k^{z}=s_k^{3d}(K_{cam}[u,v,z])$. And the 3D offsets of category $k$ can be calculated by $\triangle c_k^{z} = \triangle c^{z} * s_k^{z}$.

Meanwhile, we can get 2D instance centers $c^{2d}$ from 2D center map $I_c^{2d}$, and then the instance centers of category $k$, $c_k^{2d}$, can be indexed from $c^{2d}$. Finally, 2D instance centers and 3D offsets of category $k$ are combined to group 3D instance at multi-plane:
\vspace{-3pt}
\begin{equation}
\vspace{-1pt}
\small
   i^{z}_k(u,v) = argmin_{k_j} \|c^{2d}_{k_j} - (u+\triangle c^z_k(u,v)_u,v+\triangle c^z_k(u,v)_v) \|,
\end{equation}
where $c^{2d}_{k_j} \in \mathbb{R}^2$ is the $j$th 2D instance center of category $k$. $i^{z}_k$ is the predicted instance id at depth $z$. The 3D instance id of category $k$ at location $(u,v,z)$ can be calculated by
$i^{3d}_k(u,v,z)=i^{z}_k(u,v)(K_{cam}^{-1}[u,v,z])$.


Combining the stuff from 3D semantics, and the 3D instances grouped by our instance grouping block, we finally predict the panoptic 3D scene reconstruction results from a single image.

\subsection{Loss for BUOL}
The total loss for the proposed BUOL contains 2D loss and 3D loss. The 2D priors training loss is defined as follows:
\vspace{-3pt}
\begin{equation}
\vspace{-3pt}
\mathcal{L}^{2d}=w_p^{2d}\mathcal{L}^{2d}_p+w_d^{2d}\mathcal{L}^{2d}_d+w_o^{mp}\mathcal{L}^{mp}_o
\end{equation}
where weights $w_p^{2d}$,$w_d^{2d}$ and $w_o^{mp}$ are used to balance the objective. The panoptic segmentation loss is
\vspace{-3pt}
\begin{equation}
\vspace{-3pt}
\begin{aligned}
    \mathcal{L}^{2d}_p=w_s^{2d} CE(s^{2d}, \hat{s}^{2d})+w_c^{2d} MSE(I_c^{2d}, \hat{I}_c^{2d})\\
\end{aligned}
\end{equation}
which is composed of semantic map \textit{cross entropy} loss and instance center regression \textit{mean-squared} loss.
The ground truth center map $\hat{I}_c^{2d}$ are defined as 2D Gaussian-encoded heatmaps centered in instance mass, and the ground truth of 2D instances and 2D semantics are rendered by 3D instances and 3D semantics, respectively.
The depth estimation loss $\mathcal{L}^{2d}_d$ follows \cite{hu2019revisiting} to penalize the difference between the estimated depth $d$ and the ground truth depth $\hat{d}$ which is generated by the 3D geometry.
The multi-plane occupancy loss $\mathcal{L}^{mp}_o$ is defined as:
\vspace{-3pt}
\begin{equation}
\vspace{-3pt}
\begin{split}
    \mathcal{L}^{mp}_o=BCE(o^{mp}, \hat{o}^{mp}), 
\end{split}
\end{equation}
where the $\hat{o}^{mp}$ is obtained by sampling the 3D ground truth occupancy $\hat{o}^{3d}$ at multi-plane, \ie $\hat{o}^{mp}=\hat{o}^{3d}(K_{cam}[u,v,z]) $.

The 3D loss of BUOL is composed of 3D occupancy loss, 3D semantic loss, and 3D offset loss, defined as follows:
\vspace{-3pt}
\begin{equation}
\vspace{-3pt}
\begin{split}
    \mathcal{L}^{3d}=w_o^{3d}\mathcal{L}^{3d}_o(o^{3d},\hat{o}^{3d})    +w_s^{3d}CE(s^{3d'},\hat{s}^{3d})\\
    +w_{\triangle c}^{3d}L1(\triangle c^{3d'},\triangle \hat{c}^{3d})
\end{split}
\end{equation}
where $w_o^{3d}, w_s^{3d}, w_{\triangle c}^{3d}$ are weighting coefficients. The 3D occupancy loss $\mathcal{L}^{3d}_o$ is composed of a binary classification loss $BCE$ and a regression loss $L1$, and the details can be referred to supplemental materials.
The ground truth $\triangle \hat{c}^{3d}$ for each voxel is offset between its 2D instance center and location in its nearest depth plane, which can be generated by 3D ground truth instances. 

To stabilize the training, we first train 2D model $F_{\theta}$ with $\mathcal{L}^{2d}$. After converging, the 3D loss $\mathcal{L}^{3d}$ is applied to train 3D model $G_{\phi}$.

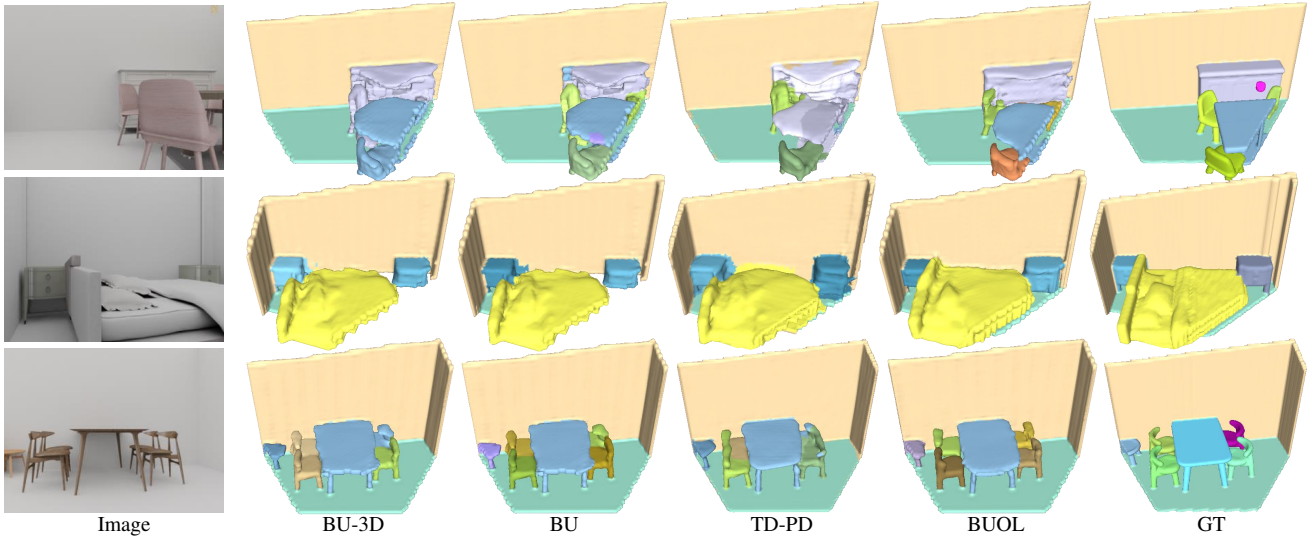
\begin{figure*}[t!]
\centering
\small
\setlength\tabcolsep{0pt}
\includegraphics[width=1.0\linewidth]{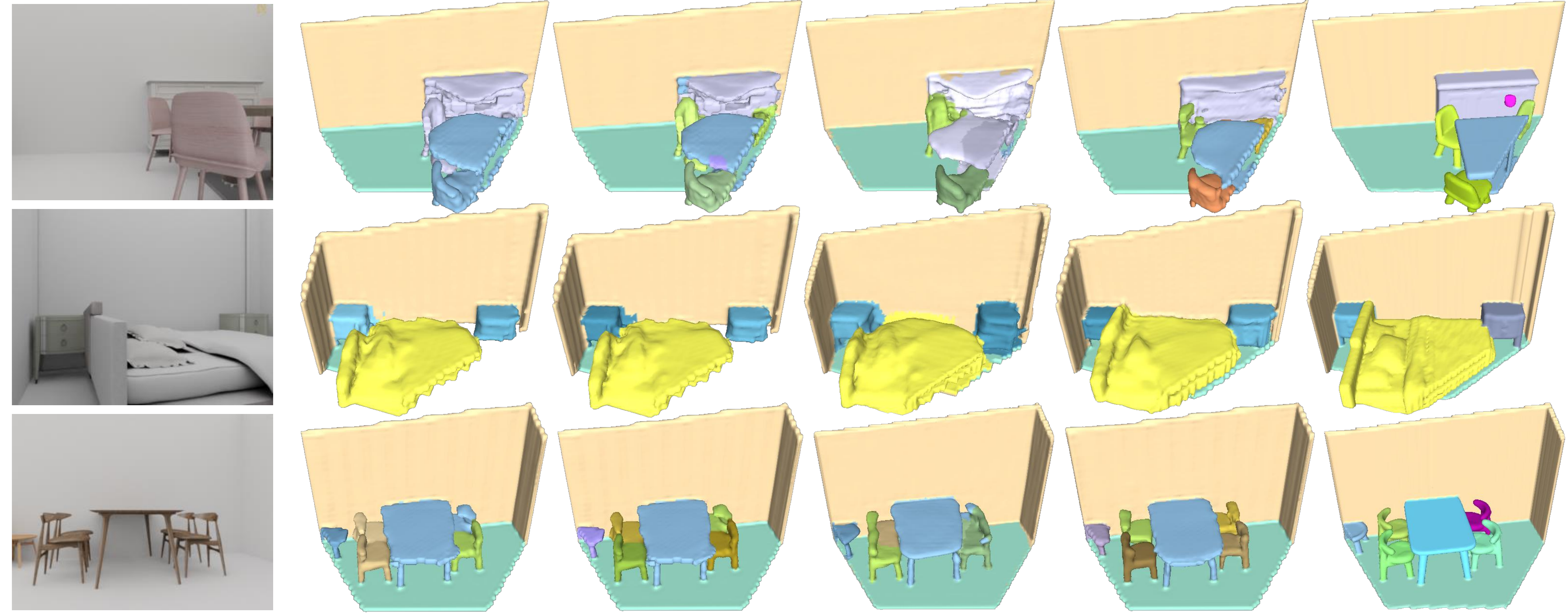}
\put(-458,-6){\footnotesize Image}
\put(-373,-6){\footnotesize BU-3D}
\put(-287,-6){\footnotesize BU}
\put(-211,-6){\footnotesize TD-PD}
\put(-129,-6){\footnotesize BUOL}
\put(-42,-6){\footnotesize GT}
\caption{\textbf{Qualitative comparisons against competing methods on 3D-Front.} The BUOL and BU denote our Bottom-Up framework w/ and w/o our Occupancy-aware lifting, respectively, and BU-3D denotes the bottom-up framework with instance grouping by 3D centers, and the TD-PD denotes Dahnert et al. \cite{dahnert2021panoptic}$^*$+PD. And GT is the ground truth.}
\label{fig:visualization}
\end{figure*}

\begin{table*}[t]
\centering
\begin{tabular}{l|ccc|ccc|ccc}
    Method & PRQ & RSQ & RRQ & PRQ$_\text{th}$ & RSQ$_\text{th}$ & RRQ$_\text{th}$ & PRQ$_\text{st}$ & RSQ$_\text{st}$ & RRQ$_\text{st}$\\
    \hline
    SSCNet \cite{song2017semantic}+IC & 11.50 & 32.90 & 33.00 & 8.03 & 32.07 & 24.69 & 26.95 & 36.75 & 70.25 \\
    Mesh R-CNN \cite{gkioxari2019mesh} & - & - & - & 20.90 & 38.00 & 53.20 & - & - & - \\
    Total3D \cite{nie2020total3dunderstanding} & 15.08 & 36.63 & 40.15 & 13.77 & 34.88 & 38.89 & 20.94 & 44.49 & 45.85 \\
    Dahnert et al. \cite{dahnert2021panoptic}$^*$ & 42.20 & 55.59 & 73.19 & 36.51 & 51.47 & 69.21 & 67.78 & 74.15 & 91.09 \\
    \hline
    Dahnert et al. \cite{dahnert2021panoptic}$^*$+PD & 47.46 & 60.48 & 76.09 & 42.25 & 56.90 & 72.45 & 70.94 & 76.59 & 92.45 \\
    Our BUOL & \textbf{54.01} & \textbf{63.81} & \textbf{82.99} & \textbf{49.73} & \textbf{60.57} & \textbf{80.67} & \textbf{73.30} & \textbf{78.37} & \textbf{93.42} \\
    
\end{tabular}
\caption{Comparison to the state-of-the-art on 3D-Front. ``*" denotes the trained model with the official codebase released by the authors.}
\label{tab:sota}
\end{table*}

\section{Experiments}
\label{sec:experiments}
In this section, we conduct experiments on the pre-processed synthetic dataset 3D-Front \cite{fu20213d} and real-world dataset Matterport3D \cite{chang2017matterport3d}. We compare our method with state-of-the-art panoptic 3D scene reconstruction methods and provide an ablation study to highlight the effectiveness of each component.
\subsection{Experiment Setup}

\noindent\textbf{Datasets.} 3D-Front \cite{fu20213d} is a synthetic indoor dataset with 18,797 room scenes and 11 categories (9 for things, and 2 for stuff) in 6,801 mid-size apartments. To generate data for panoptic 3D scene reconstruction, we follow Dahnert et al. \cite{dahnert2021panoptic}, and first randomly sample rooms and camera locations, then use BlenderProc~\cite{denninger2019blenderproc} to render RGB images along with depth, semantic map, and instance mask and finally use signed distance function (SDF) to get 3D ground truth.
It contains 96,252/11,204/26,933 train/val/test images corresponding to 4,389/489/1,206 scenes, respectively.


Matterport3D \cite{chang2017matterport3d} is a real-world indoor dataset that contains 90 building-scale scenes. For panoptic 3D scene reconstruction, Matterport3D is pre-processed in the same way as 3D-Front to generate the ground truth of 34,737/4,898/8,631 train/val/test images corresponding to 61/11/18 scenes. It contains the same 11 categories as 3D-Front and another stuff category ``ceiling".


\noindent\textbf{Metrics.} We adopt panoptic reconstruction quality $PRQ$, reconstructed segmentation quality $RSQ$, and reconstructed recognition quality $RRQ$~\cite{dahnert2021panoptic} as our metrics. In addition, $PRQ_\text{th}$ and $PRQ_\text{st}$ denote $PRQ$ of things and stuff, respectively. $PRQ$ is calculated by the average measure across $C$ categories, with $PRQ_k$ for category $k$ defined as:
\vspace{-5pt}
\begin{equation}
\small
\begin{aligned}
    PRQ_k & =RSQ_k*RRQ_k\qquad \\
    &=\frac{\sum_{(i,\hat{i})\in TP_k}IoU(i,\hat{i})}{|TP_k|}*\frac{2|TP_k|}{2|TP_k|+|FP_k|+|FN_k|}\\
    &=\frac{\sum_{(i,\hat{i})\in TP_k} 2 IoU(i,\hat{i})}{2|TP_k|+|FP_k|+|FN_k|}
\end{aligned}
\end{equation}
where $TP_k$, $FP_k$, and $FN_k$ denote true positives, false positives, and false negatives for category $k$, respectively, and intersection over union ($IoU$) is the metric between predicted mask $i$ and ground truth mask $\hat{i}$. The predicted segments are matched with ground truth if the voxelized $IoU$ is no less than $25\%$. Following Dahnert et al. \cite{dahnert2021panoptic}, we set the evaluate resolution for panoptic 3D scene reconstruction to 3cm for synthetic data and 6cm for real-world data.

\begin{figure*}[t!]
\centering
\small
\setlength\tabcolsep{0pt}
\includegraphics[width=1.0\linewidth]{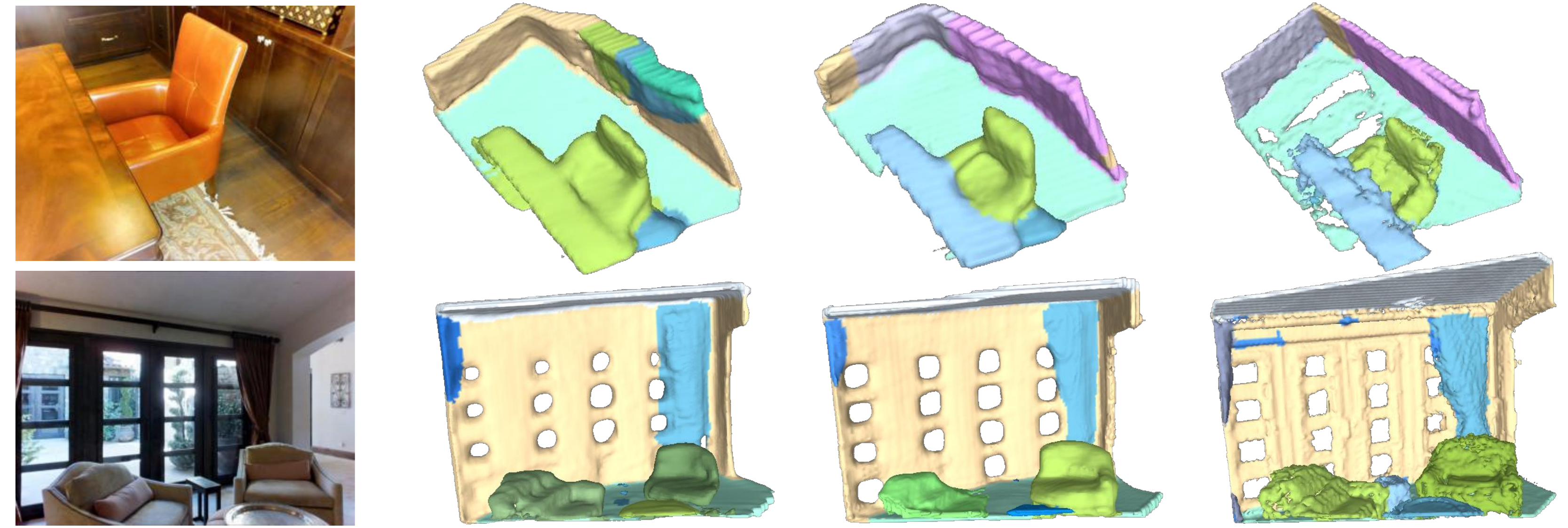}
\put(-447,-6){\footnotesize Image}
\put(-316,-6){\footnotesize TD-PD}
\put(-192,-6){\footnotesize BUOL}
\put(-65,-6){\footnotesize GT}
\caption{\textbf{Qualitative comparisons against competing methods on Matterport3D.} The BUOL denotes our Bottom-Up framework with Occupancy-aware lifting, and the ``TD-PD" denotes Dahnert et al. \cite{dahnert2021panoptic}$^*$+PD. And GT is the ground truth.}
\label{fig:visualization_m3d}
\end{figure*}
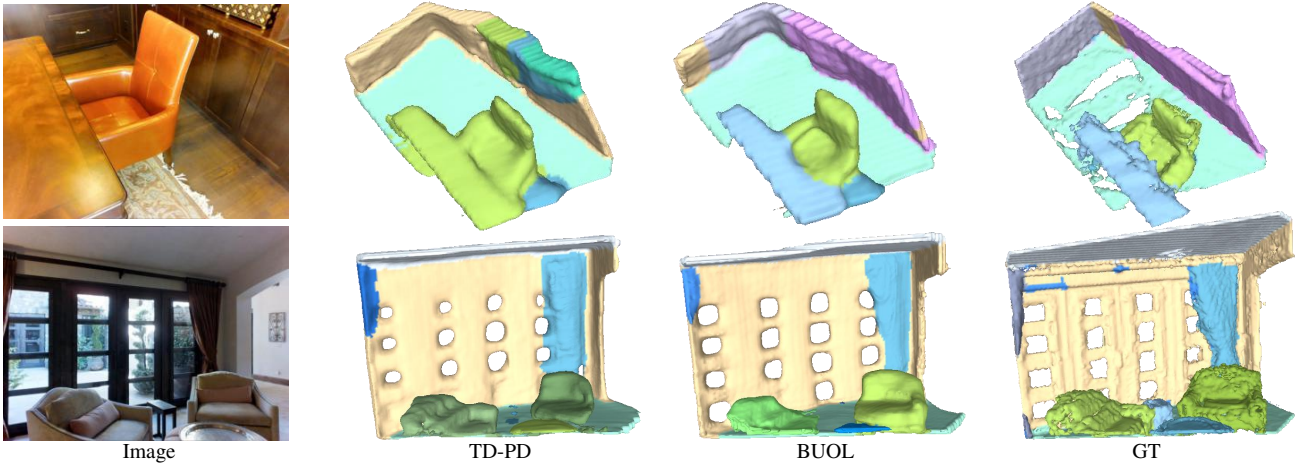

\begin{table*}[t]
\centering
\begin{tabular}{l|ccc|ccc|ccc}
    Method & PRQ & RSQ & RRQ & PRQ$_\text{th}$ & RSQ$_\text{th}$ & RRQ$_\text{th}$ & PRQ$_\text{st}$ & RSQ$_\text{st}$ & RRQ$_\text{st}$\\
    \hline
    SSCNet \cite{song2017semantic}+IC & 0.49 & 21.68 & 1.50 & 0.19 & 22.75 & 0.59 & 1.43 & 20.43 & 4.43 \\
    Mesh R-CNN \cite{gkioxari2019mesh} & - & - & - & 6.29 & 31.12 & 15.60 & - & - & - \\
    Dahnert et al. \cite{dahnert2021panoptic} & 7.01 & 28.57 & 17.65 & 6.34 & 26.06 & 16.06 & 10.78 & 40.03 & 26.77 \\
    \hline
    Dahnert et al. \cite{dahnert2021panoptic}$^*$+PD & 10.08 & 36.04 & 22.53 & 7.33 & 33.23 & 16.68 & 18.33 & 44.47 & 40.07 \\
    Our BUOL & \textbf{14.47} & \textbf{45.71} & \textbf{30.91} & \textbf{10.97} & \textbf{45.30} & \textbf{23.81} & \textbf{24.94} & \textbf{46.93} & \textbf{52.22} \\
    
\end{tabular}
\caption{Comparison to the state-of-the-art on Matterport3D. ``*" denotes the trained model with the official codebase released by the authors.}
\label{tab:sota_mp}
\end{table*}

\noindent\textbf{Implementation.} We adopt ResNet-50~\cite{he2016deep} as our shared 2D backbone of 2D Panoptic-Deeplab~\cite{cheng2020panoptic}, and use three branches to learn rich 2D priors. One decoder with the semantic head is used for semantic segmentation, and one decoder followed by the center head
is utilized for instance
center estimation. Another decoder with a depth head and multi-plane occupancy head is designed for geometry priors.
For the 3D model, we convert 2D ResNet-18~\cite{he2016deep} and ASPP-decoder~\cite{chen2017deeplab} to 3D models as our 3D encoder-decoder, and design 3D occupancy head, 3D semantic head, and 3D offset head for panoptic 3D scene reconstruction. For the two datasets, we apply Adam~\cite{kingma2014adam} solver with the initial learning rate 1e-4 combined with polynomial learning rate decay scheduler for 2D learning, and the initial learning rate 5e-4 decayed at 32,000th and 38,000th iteration. During training, we first train the 2D model for 50,000 iterations with batch size 32, then freeze the parameters and train the 3D model for 40,000 iterations with batch size 8. All the experiments are conducted with 4 Tesla V100 GPUs. In addition, we initialize the model with the pre-trained ResNet-50 for 3D-Front, and the pre-trained model on 3D-Front for Matterport3D which is the same as Dahnert et al. \cite{dahnert2021panoptic}.


\subsection{Comparison with State-of-the-art Methods}

\textbf{3D-Front.} For synthetic dataset, we compare BUOL with the state-of-the-art method~\cite{dahnert2021panoptic} and other mainstream~\cite{song2017semantic, gkioxari2019mesh, nie2020total3dunderstanding}. The results are shown in Table~\ref{tab:sota}. Our proposed BUOL and Dahnert et al. \cite{dahnert2021panoptic} both outperform other methods a lot. However, with the proposed bottom-up framework and occupancy-aware lifting, our BUOL outperforms the state-of-the-art method by a large margin, +11.81\%. For a fair comparison, we replace the 2D segmentation in Dahnert et al. \cite{dahnert2021panoptic} with the same 2D model as ours, denoted as Dahnert et al. \cite{dahnert2021panoptic}+PD (also denoted as TD-PD). Comparing to this method in Table~\ref{tab:sota}, our BUOL also shows an advantage of +6.55\% PRQ. The qualitative comparison results in  Figure.~\ref{fig:visualization} also show our improvement. In the second row, our BUOL reconstructs the bed better than TD-PD with occupancy-aware lifting. In the last row, our BUOL can recognize all the chairs while TD-PD obtains the sticky chair. In the first rows, both BU and OL in BUOL achieve the better Panoptic 3D Scene Reconstruction results.

\textbf{Matterport3D.} We also compare BUOL with some methods \cite{dahnert2021panoptic,song2017semantic, gkioxari2019mesh} on real-world dataset. The results are shown in Table~\ref{tab:sota_mp}. Our BUOL outperforms the state-of-the-art Dahnert et al. \cite{dahnert2021panoptic} by +7.46\% PRQ with the proposed bottom-up framework and occupancy-aware lifting. For fairness, we also compare BUOL with Dahnert et al. \cite{dahnert2021panoptic}+PD, and our method improves the PRQ by +4.39\%. Figure.~\ref{fig:visualization_m3d} provides the qualitative results. In the first row, our BUOL can segment all instances corresponding to ground truth, which contains a chair, a table and two cabinets, and TD-PD can only segment the chair. In the second row, our BUOL reconstruct the wall and segment curtains batter than TD-PD. In addition, although the highest performance, the PRQ of the Matterport3D is still much lower than that of the 3D-Front due to its noisy ground truth.

\subsection{Ablation Study}
In this section, we verify the effectiveness of our BUOL for panoptic 3D scene reconstruction. As shown in Table~\ref{tab:ablation}, for a fair comparison, TD-PD is the state-of-the-art top-down method~\cite{dahnert2021panoptic} with the same 2D Panoptic-Deeplab as ours, which is our baseline method. BU denotes our proposed bottom-up framework. Different from BU, 2D Panoptic-Deeplab in TD-PD is used to predict instance masks instead of semantics and instance centers. BU-3D denotes the bottom-up framework which groups 3D instances by the predicted 3D centers instead of 2D centers.


\noindent\textbf{Top-down vs. Bottom-up.}
TD-PD and BU adopt the same 2D model. The former lifts the instance masks to the 3D features, while the latter lifts the semantic map and groups 3D instances with 2D instance centers. Comparing the two settings in Table~\ref{tab:ablation}, BU significantly boosts the performance of RRQ by +5.85\% which proves our bottom-up framework with proposed 3D instance grouping achieves more accurate 3D instance mask than direct instance mask lifting. The drop of PRQ$_\text{st}$ for stuff may come from the lower capability of used 3D ResNet + ASPP, compared with other methods equipped with stronger but memory-consuming 3D UNet. Overall, the proposed bottom-up 
framework achieves +3.3\% PRQ better than the top-down method. Figure.~\ref{fig:visualization} provides qualitative comparison of BU and TD-PD. The bottom-up framework performs better than the top-down method. For example, in the last row of Figure.~\ref{fig:visualization}, TD-PD fails to recognize the four chairs, while BU reconstructs and segments better.

\noindent\textbf{2D instance center vs. 3D instance center.}
We also compare the 2D instance center with the 3D instance center for 3D instance grouping. To estimate the 3D instance center, the center head is added to the 3D refinement model, called BU-3D. Quantitative comparing BU-3D and BU in Table~\ref{tab:ablation}, we can find the PRQ$_\text{st}$ for stuff is similar, but when grouping 3D instances with the 2D instance centers, the PRQ$_\text{th}$ for thing has improved by 5.03\%, which proves 3D instance grouping with 2D instance center performing better than that with the 3D instance center.
We conjecture that the error introduced by the estimated depth dimension may impact the position of the 3D instance center.
Meanwhile, grouping in multi-plane is easier for 3D offset learning via reducing one dimension to be predicted. Qualitative comparison BU with BU-3D is shown in Figure.~\ref{fig:visualization}, due to inaccurate 3D instance centers, the result of BU-3D in the last row misclassifies a chair as a part of the table, and the result in the first row does not recognize one chair.

\begin{table}[t]
\centering
\begin{tabular}{l|ccc|cc}
    Method & PRQ & RSQ & RRQ & PRQ$_\text{th}$ & PRQ$_\text{st}$\\
    \hline
    TD-PD & 47.46 & 60.48 & 76.09 & 42.25 & 70.94 \\
    BU-3D & 46.73 & 59.17 & 76.68 & 41.77 & 69.07 \\
    \hline
    BU & 50.76 & 60.66 & 81.94 & 46.80 & 68.55 \\
    BUOL & \textbf{54.01} & \textbf{63.81} & \textbf{82.99} & \textbf{49.73} & \textbf{73.30} \\
\end{tabular}
\caption{Ablation study of the proposed method vs baselines.}
\label{tab:ablation}
\end{table}

\begin{table}[t]
\centering
\begin{tabular}{c|c|ccc}
    Inst/Sem & Assignment & PRQ & RSQ & RRQ\\
    \hline
    \multirow{2}{*}{Instance} & random & 47.46 & 60.48 & 76.09 \\
    & category & 48.92 & 61.20  & 77.48 \\
    \cline{1-5}
    Semantics & category & 50.76 & 60.66 & 81.94 \\
\end{tabular}
\caption{Comparison of different assignments.
}
\label{tab:random}
\end{table}

\noindent\textbf{Voxel-reconstruction ambiguity.} 
We propose occupancy-aware lifting to provide the 3D features in full 3D space to tackle voxel-reconstruction ambiguity. Quantitative comparing BUOL with BU in Table~\ref{tab:ablation}, our proposed occupancy-aware lifting improves PRQ$_\text{th}$  by 2.93\% for thing and PRQ$_\text{st}$ by 4.75\% for stuff,
which verifies the effectiveness of multi-plane occupancy predicted by the 2D model. It facilitates the 3D model to predict more accurate occupancy of the 3D scene. In addition, with our occupancy-aware lifting, PRQ$_\text{st}$ for stuff of the 3D model with ResNet-18 + 3D ASPP outperforms the model TD-PD with 3D U-Net by 2.36\% PRQ. As shown in Figure.~\ref{fig:visualization}, with occupancy-aware lifting, BUOL reconstructs 3D instances better than others. For example, in the second row of Figure.~\ref{fig:visualization},  BUOL can reconstruct the occluded region of the bed, while other settings fail to tackle this problem.

\noindent\textbf{Instance-channel ambiguity.}
To analyze the instance-channel ambiguity in the top-down method, we conduct experiments based on TD-PD, as shown in Table~\ref{tab:random}. When lifting instance masks with random assignment, the model achieves 47.76\% PRQ. However, fitting random instance-channel assignment makes the model pay less attention to scene understanding. To reduce the randomness, we try to apply instance-channel with sorted categories, which improves PRQ to 48.92\%. Because an arbitrary number of instances with different categories may exist in an image, resulting in the randomness of instance number even for the same category. To further reduce the randomness, our proposed bottom-up method, also called BU, lifts semantics with deterministic assignment, and gets 50.76\% PRQ, which proves that the pressure of the 3D model can be reduced with the reduction in the randomness of instance-channel assignment and the bottom-up method can address the instance-channel ambiguity.


\section{Conclusion}
In this paper, we propose a bottom-up framework with occupancy-aware lifting (BUOL) for panoptic 3D scene reconstruction. Our bottom-up framework lifts 2D semantics instead of 2D instances to 3D to avoid instance-channel ambiguity, and the proposed occupancy-aware lifting leverages multi-plane occupancy predicted by 2D model to avoid voxel-reconstruction ambiguity. BUOL outperforms state-of-art approaches with top-down framework for both 3D reconstruction and 3D perception in a series of experiments. We believe that BUOL will drive the area of panoptic 3D scene reconstruction from a single image forward.


This project is funded in part by the National Natural Science Foundation of China (No.61976094), Shanghai AI Laboratory, CUHK Interdisciplinary AI Research Institute, the Centre for Perceptual and Interactive Intelligence (CPIl) Ltd under the Innovation and Technology Commission (ITC)'s InnoHK, and Hong Kong RGC Theme-based Research Scheme 2020/21 (No. T41-603/20- R).

{\small
\bibliographystyle{ieee_fullname}
\bibliography{egbib}
}

\renewcommand\thesection{\Alph{section}} 
\setcounter{section}{0}
\clearpage

\section{Architecture}
\begin{figure*}[t!]
\centering
\includegraphics[width=0.8\linewidth]{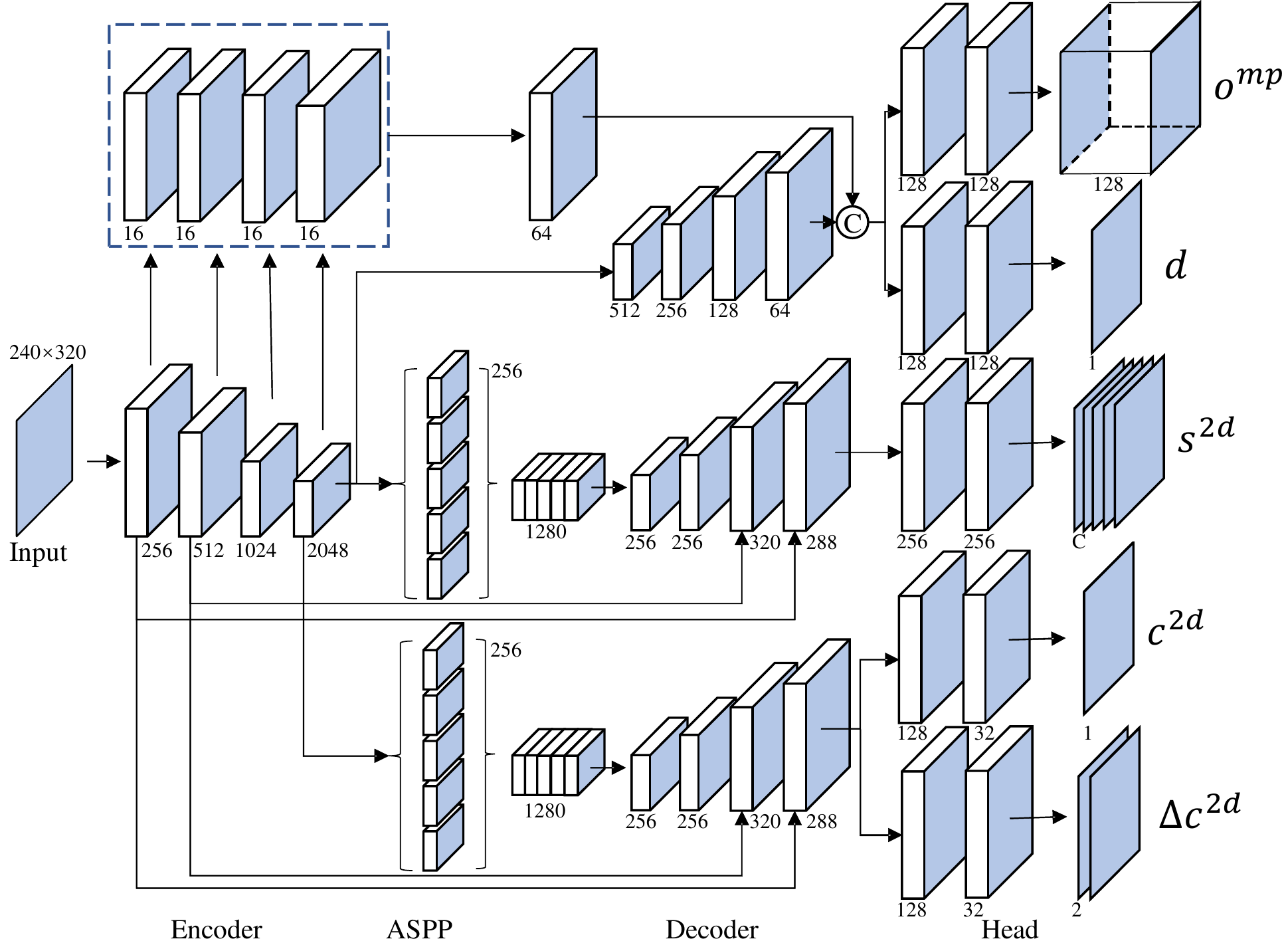}
\caption{\textbf{The 2D model in detail of our BUOL.}}
\label{fig:model_2d}
\end{figure*}

We provide the details of the 2D model and 3D model in our framework. The 2D model is shown in Figure.~\ref{fig:model_2d}, which is composed of ResNet-50~\cite{he2016deep} as a shared encoder, three decoders, and multiple prediction heads. We employ  Panoptic-Deeplab~\cite{cheng2020panoptic} as our 2D panoptic segmentation, which contains one ASPP-decoder~\cite{chen2017deeplab} followed by center head and offset head to predict 2D center map $c^{2d}$ and regress 2D offsets $\triangle c^{2d}$, respectively, and another ASPP-decoder with semantic head to predict semantic map $s^{2d}$. The final decoder followed by the depth head and multi-plane occupancy head is designed to predict depth $d$ and multi-plane occupancy $o^{mp}$, respectively.

\begin{figure*}[t!]
\centering
\includegraphics[width=1.\linewidth]{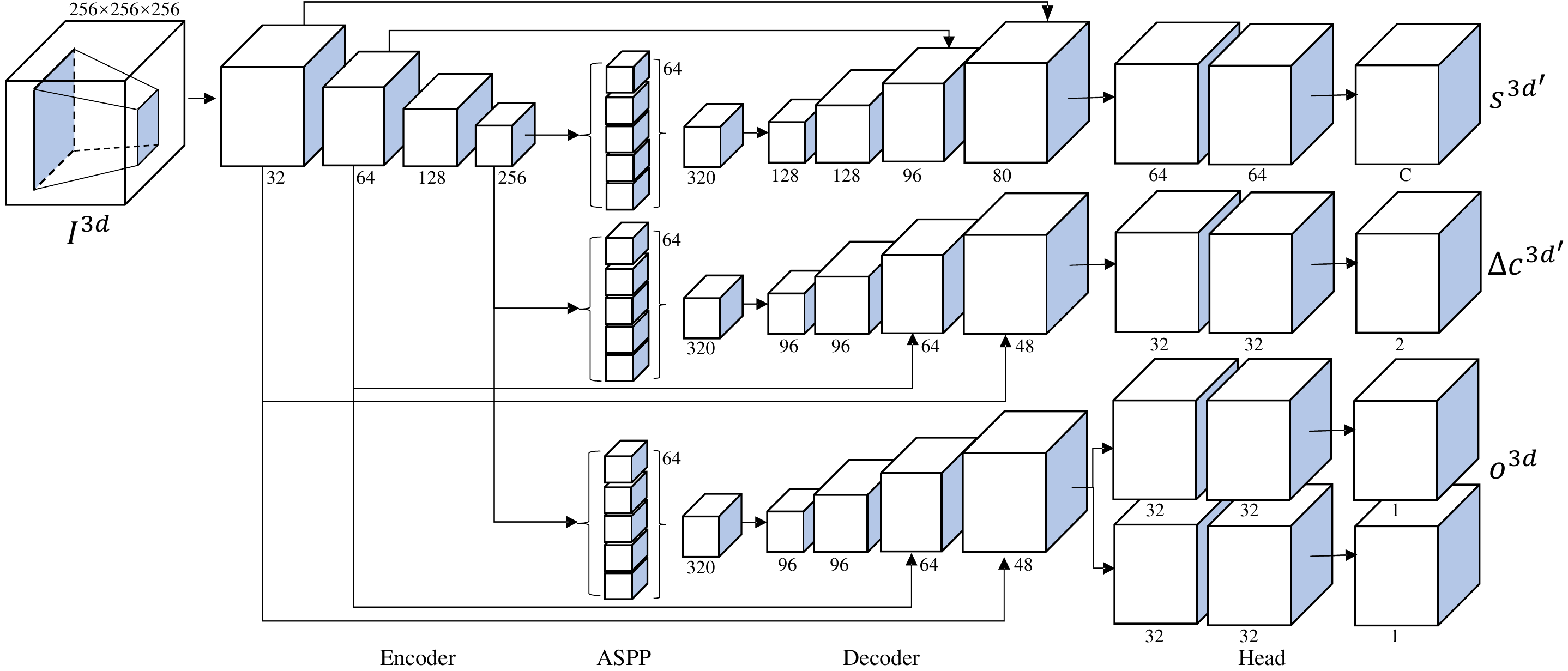}
\caption{\textbf{The 3D model in detail of our BUOL.}}
\label{fig:model_3d}
\end{figure*}

With the 3D feature generated by our occupancy-aware lifting as input, a 3D refinement model is applied to predict 3D results
, as shown in Figure.~\ref{fig:model_3d}. Specifically, we convert ResNet-18~\cite{he2016deep} and ASPP-decoder~\cite{chen2017deeplab} to 3D to combine into our 3D encoder-decoder and reduce the channels of the 3D model due to limitations on graphics memory. Similar to the 2D model, we adapt one shared encoder and three decoders for 3D prediction. 3D semantic map $s^{3d'}$ is refined by the 3D semantic head following one decoder, and 3D offsets $\triangle c^{3d'}$ is predicted by 3D offset head following another decoder, and 3D occupancy $o^{3d}$ is obtained by the dot product of two outputs predicted by two heads with the last shared decoder.
One output with $BCE$ loss is designed to obtain the course 3D occupancy, and the other with $L1$ loss regresses the Truncated Signed Distance Function (TSDF).
Finally, The panoptic 3D scene reconstruction result is processed by our proposed bottom-up panoptic reconstruction.

\section{Additional Quantitative Results}
Table~\ref{tab:prq_class} shows the PRQ for each category on the synthetic dataset 3D-Front. Our bottom-up framework ``BU" outperforms the top-down methods~\cite{dahnert2021panoptic,song2017semantic, gkioxari2019mesh, nie2020total3dunderstanding} for almost all categories. With our occupancy-aware lifting, ``BUOL" achieves further better performance for each category. Overall, our BUOL improves the baseline by +6.55\% PRQ and the state-of-the-art~\cite{dahnert2021panoptic} by +11.81\% PRQ, respectively.

The PRQ for each category on the real-world dataset Matterport3D is shown in Table~\ref{tab:prq_class_m3d}. Our proposed BUOL outperforms the other top-down methods~\cite{dahnert2021panoptic,song2017semantic, gkioxari2019mesh} a lot.
Compared with Dahnert et al. \cite{dahnert2021panoptic}, our method achieves higher performance for all categories and improves the overall PRQ by 7.46\%. For a fair comparison, we also compare BUOL with our strong baseline Dahnert et al. \cite{dahnert2021panoptic}+PD, and our framework achieves +4.39\% PRQ better. The quantitative results show that our bottom-up framework with occupancy-aware lifting outperforms the state-of-the-art methods in both synthetic and real-world datasets.


\section{Additional Qualitative Results}
\begin{figure*}[t!]
\centering
\small
\setlength\tabcolsep{0pt}
\includegraphics[width=1.0\linewidth]{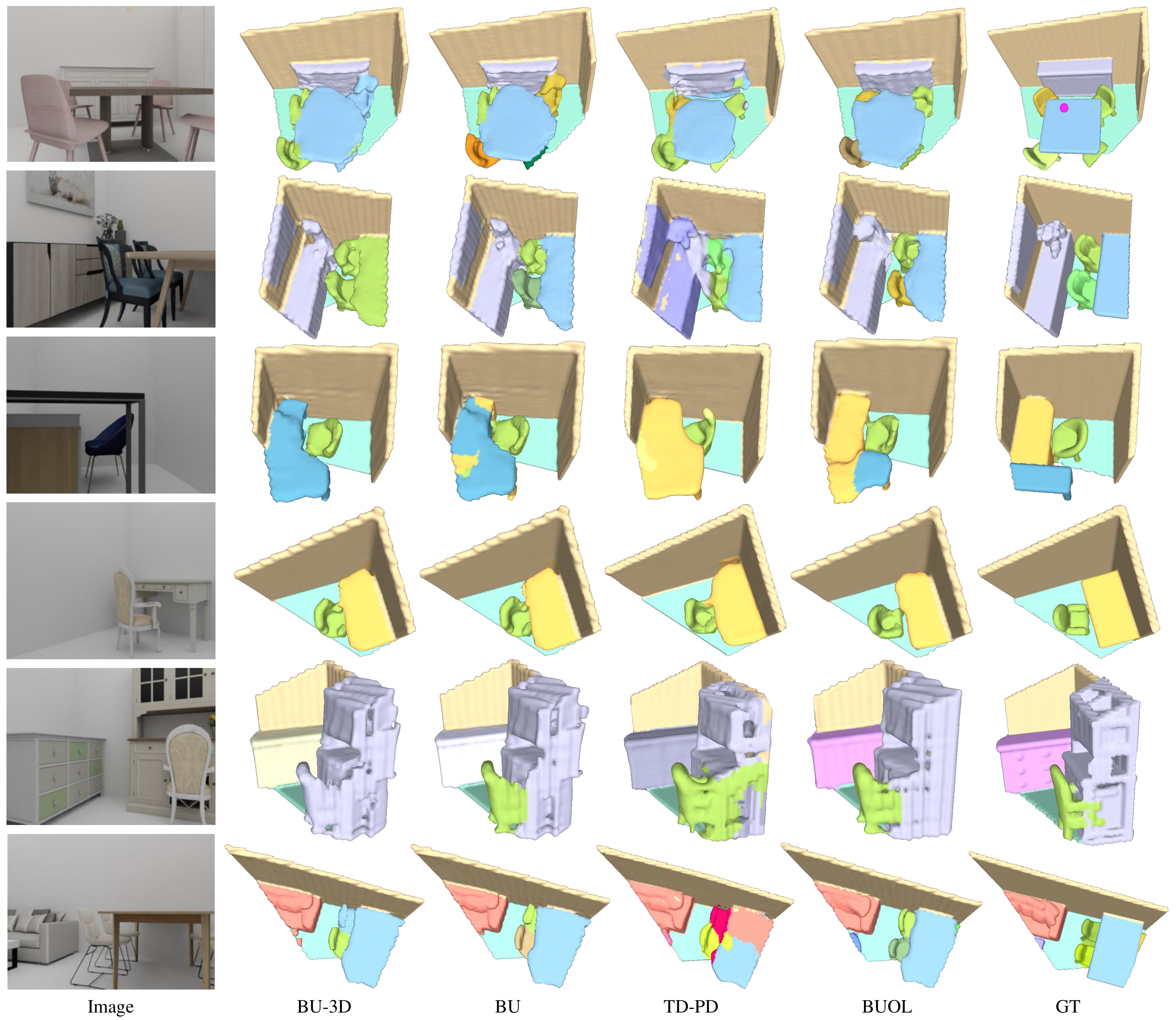}
\caption{\textbf{Qualitative comparisons on 3D-Front.} The BUOL and BU denote our Bottom-Up framework w/ and w/o our Occupancy-aware lifting, respectively, and BU-3D denotes the bottom-up framework with instance grouping by 3D centers, and the TD-PD denotes Dahnert et al. \cite{dahnert2021panoptic}$^*$+PD. And GT is the ground truth.}
\label{fig:supp_vis}
\end{figure*}

Our additional qualitative results on the synthetic dataset 3D-Front are shown in Figure.~\ref{fig:supp_vis}. Comparing the results in each row, our BUOL reconstructs instances and segments them better. For example, the chairs in rows 1, 2, 5, and 6 are more complete than the other models, and the shape of instances in each row is also closer to the ground truth.

Figure.~\ref{fig:supp_vis_m3d} shows the additional qualitative results on the real-world dataset Matterport3D. Both thing and stuff categories are reconstructed and segmented better by our BUOL. Comparing the shape of the wall in each row, our method performs closer to the ground truth. And comparing the model results of the pillows in row 1, the chairs in rows 3 (upper left) and 5 (left), and the table in row 3, our BUOL assigns the correct category to each instance while TD-PD assigns the wrong category. And comparing the model results of the flowers and platform in row 2, the floor and instances in row 4, and the chairs in row 5, the reconstruction results of our method are better.

Comparing all the results in detail of both synthetic and real-world datasets, both ``BU" and ``OL" in our Bottom-Up framework with Occupancy-aware Lifting lead to better Panoptic 3D Scene Reconstruction from a single image.

\clearpage

\begin{table*}[t]
\centering
\setlength{\tabcolsep}{1.6mm}{
\begin{tabular}{c|ccccccccc|cc|c}
    Method & Cabinet & Bed & Chair & Sofa & Table & Desk & Dresser & Lamp & Other & Wall & Floor & PRQ \\
    \hline
    SSCNet \cite{song2017semantic}+IC & 7.80 & 16.60 & 7.90 & 13.30 & 12.10 & 5.50 & 0.50 & 0.70 & 7.90 & 15.20 & 38.70 & 11.50 \\
    Mesh R-CNN \cite{gkioxari2019mesh} & 29.70 & 13.30 & 24.10 & 24.40 & 28.50 & 23.50 & 14.40 & 1.40 & 28.70 & - & - & -\\
    Total3D \cite{nie2020total3dunderstanding} & 17.25 & 4.56 & 18.76 & 14.07 & 19.40 & 16.79 & 7.04 & 8.13 & 17.97 & 8.27 & 33.61 & 15.08 \\
    Dahnert et al. \cite{dahnert2021panoptic}$^*$ & 43.59 & 49.86 & 27.43 & 42.07 & 40.18 & 34.05 & 40.43 & 8.77 & 42.21 & 57.63 & 77.93 & 42.20 \\
    \hline
    Dahnert et al. \cite{dahnert2021panoptic}$^*$+PD & 47.67 & 59.00 & 36.59 & 51.96 & 42.72 & 38.70 & 47.18 & 12.47 & 43.93 & 62.22 & 79.66 & 47.46 \\
    Our BU & 54.81 & 55.77 & 43.10 & 54.81 & 50.60 & 49.02 & 49.23 & 14.74 & 49.15 & 62.73 & 74.37 & 50.76 \\
    Our BUOL & \textbf{56.06} & \textbf{64.13} & \textbf{46.46} & \textbf{56.61} & \textbf{52.72} & \textbf{52.08} & \textbf{50.97} & \textbf{17.44} & \textbf{51.08} & \textbf{64.63} & \textbf{81.97} & \textbf{54.01} \\
    
\end{tabular}}
\caption{The PRQ for each category on 3D-Front. ``*" denotes the trained model with the official codebase released by the authors.}
\label{tab:prq_class}
\end{table*}

\begin{figure*}[t!]
\centering
\small
\setlength\tabcolsep{0pt}
\includegraphics[width=1.0\linewidth]{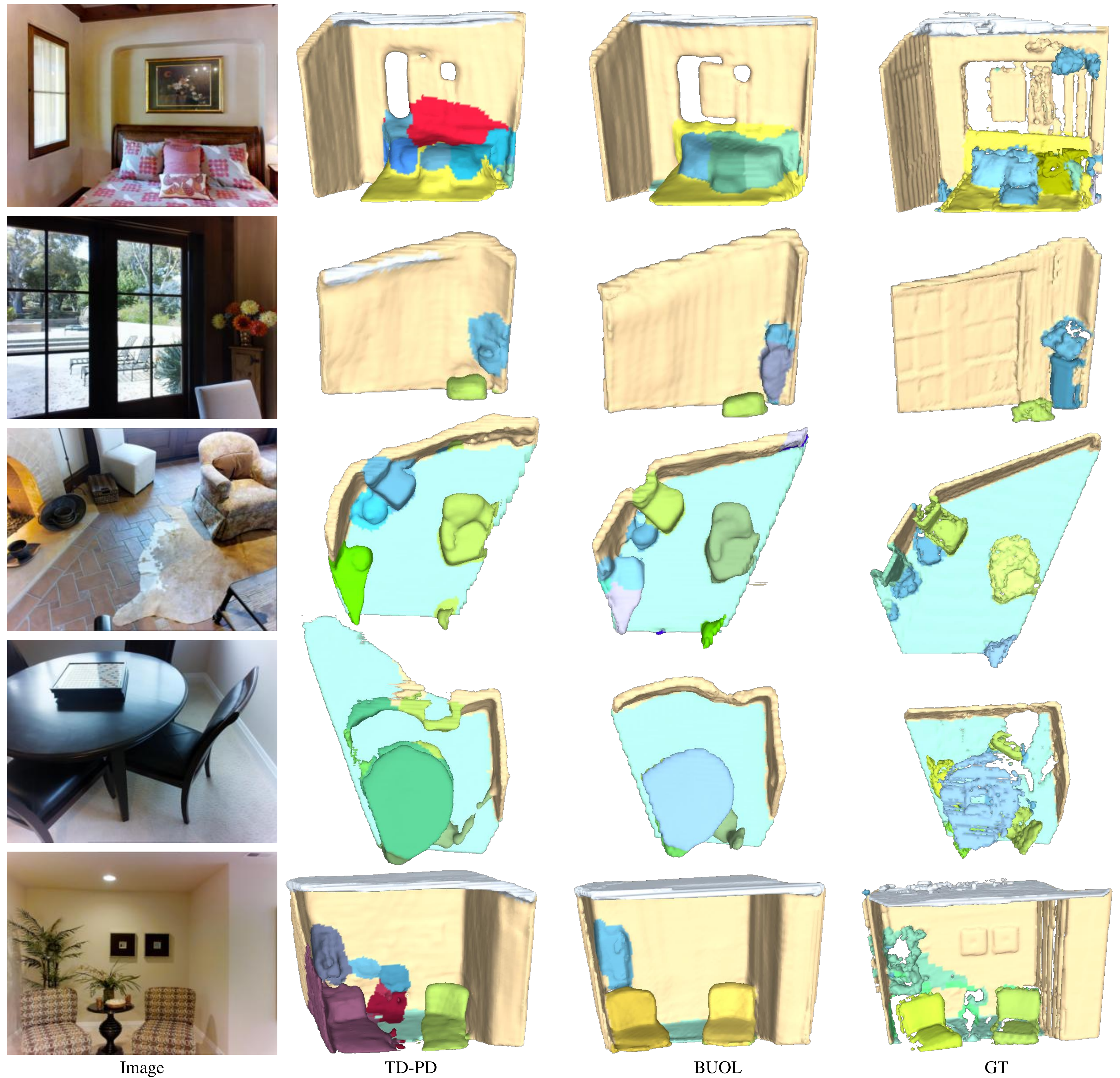}
\caption{\textbf{Qualitative comparisons on Matterport3D.} The BUOL denotes our Bottom-Up framework with Occupancy-aware lifting, and the ``TD-PD" denotes Dahnert et al. \cite{dahnert2021panoptic}$^*$+PD. And GT is the ground truth.}
\label{fig:supp_vis_m3d}
\end{figure*}

\begin{table*}[t]
\centering
\setlength{\tabcolsep}{1.1mm}{
\begin{tabular}{c|ccccccccc|ccc|c}
    Method & Cabinet & Bed & Chair & Sofa & Table & Desk & Dresser & Lamp & Other & Wall & Floor & Ceiling & PRQ \\
    \hline
    SSCNet \cite{song2017semantic}+IC & 0.07 & 0.11 & 0.61 & 0.07 & 0.53 & 0.00 & 0.00 & 0.00 & 0.19 & 0.34 & 3.96 & 0.00 & 0.49 \\
    Mesh R-CNN \cite{gkioxari2019mesh} & 3.10 & 10.00 & 14.80 & 12.00 & 7.90 & 0.00 & 0.00 & 2.80 & \textbf{6.00} & - & - & - & -\\
    Dahnert et al. \cite{dahnert2021panoptic} & 12.33 & 10.24 & 9.75 & 14.40 & 8.07 & 0.00 & 0.00 & 0.00 & 2.26 & 10.92 & 16.54 & 4.88 & 7.01 \\
    \hline
    Dahnert et al. \cite{dahnert2021panoptic}$^*$+PD & 9.73 & 20.13 & 11.95 & 12.19 & 4.87 & 0.00 & 0.00 & 2.68 & 4.42 & 16.72 & 31.53 & 6.73 & 10.08\\
    Our BUOL & \textbf{13.24} & \textbf{27.67} & \textbf{16.26} & \textbf{17.88} & \textbf{11.68} & \textbf{1.21} & \textbf{1.52} & \textbf{3.58} & 5.73 & \textbf{19.97} & \textbf{38.26} & \textbf{16.59} & \textbf{14.47} \\
    
\end{tabular}}
\caption{The PRQ for each category on Matterport3D. ``*" denotes the trained model with the official codebase released by the authors.}
\label{tab:prq_class_m3d}
\end{table*}

\end{document}